\pdfoutput=1

\documentclass[11pt]{article}
\DeclareUnicodeCharacter{FF0C}{,}
\usepackage[final]{acl}
\usepackage{times}
\usepackage{latexsym}
\usepackage{adjustbox}

\usepackage{xspace}
\newcommand{\framework}{DatawiseAgent\xspace}

\usepackage{booktabs,multirow,longtable,makecell} 
\usepackage{amsmath,amsfonts,bm}

\def\eqref#1{equation~\ref{#1}}

\def\1{\bm{1}}

\DeclareMathAlphabet{\mathsfit}{\encodingdefault}{\sfdefault}{m}{sl}
\SetMathAlphabet{\mathsfit}{bold}{\encodingdefault}{\sfdefault}{bx}{n}

\usepackage{algorithm}
\usepackage{algorithmic}
\usepackage{amsmath}
\usepackage[capitalize]{cleveref}
\usepackage{comment}
\usepackage{colortbl} 
\usepackage{caption} 
\usepackage{listings}
\usepackage{xcolor} 

\usepackage{color,frame,tcolorbox}
\usepackage{subcaption}
\usepackage{xcolor}
\usepackage[T1]{fontenc}

\usepackage[utf8]{inputenc}
\usepackage{hyperref}
\usepackage{microtype}

\usepackage{inconsolata}
\usepackage{makecell}
\usepackage{graphicx}

\title{DatawiseAgent: A Notebook-Centric LLM Agent Framework for Adaptive and Robust Data Science Automation}

\author{
  Ziming You$^{1}$, Yumiao Zhang$^{2}$, Dexuan Xu$^{3}$, Yiwei Lou$^{3}$, \\
  \textbf{Yandong Yan}$^{3}$, \textbf{Wei Wang}$^{4}$, \textbf{Huamin Zhang}$^{5}$, \textbf{Yu Huang}$^{1}$ \Thanks{~~Corresponding author: Yu Huang} \\
$^1$ National Engineering Research Center for Software Engineering, Peking University \\ 
$^2$ School of Software \& Microelectronics, Peking University \\
$^3$ School of Computer Science, Peking University \\
$^4$ Xi’an Jiaotong University \\
$^5$ Institute of Basic Theory of Chinese Medicine, China Academy of Chinese Medical Sciences \\
  \texttt{ zimingyou@stu.pku.edu.cn，hy@pku.edu.cn} 
}

\begin{document}
\maketitle
\begin{abstract}
Existing large language model (LLM) agents for automating data science show promise, but they remain constrained by narrow task scopes, limited generalization across tasks and models, and over-reliance on state-of-the-art (SOTA) LLMs. 
We introduce \textbf{\framework} \footnote{Our data and code are open-sourced at \url{https://github.com/zimingyou01/DatawiseAgent}.}, a notebook-centric LLM agent framework for adaptive and robust data science automation. 
Inspired by how human data scientists work in computational notebooks, \framework introduces a unified interaction representation and a multi-stage architecture based on finite-state transducers (FSTs). This design enables flexible long-horizon planning, progressive solution development, and robust recovery from execution failures.
Extensive experiments across diverse data science scenarios and models show that \framework consistently achieves SOTA performance by surpassing strong baselines such as AutoGen and TaskWeaver, demonstrating superior effectiveness and adaptability. 
Further evaluations reveal graceful performance degradation under weaker or smaller models, underscoring the robustness and scalability. 
\end{abstract}

\section{Introduction}
Data science, the practice of extracting knowledge and insights from data, spans a broad spectrum of processes from data gathering and interpretation to model building and decision making~\cite{donoho201750,zhang2024benchmarking}. 
As demand for data-driven decision-making continues to grow, automating data science has become a longstanding and critical challenge.
Although traditional efforts such as AutoML~\cite{he2021automl, jin2023autokeras} have achieved success in well-defined stages such as model selection and hyperparameter tuning, 
broader tasks remain difficult to formalize or mechanize due to their inherently exploratory, interdependent, and context-independent nature~\cite{bie2022automating}.

Recent advances in Large Language Models (LLMs) and LLM-based agents~\cite{xue2023dbgpt, cheng2023gpt4, dibia2023lida, hollmann2024feature, zhang2023hyper} have opened new possibilities for automating data science.
LLMs demonstrate strong zero/few-shot generalization, in-context reasoning, code generation, and tool use capabilities, enabling a new line of research on \emph{data science agents} ~\cite{zhang2024benchmarking}, which are autonomous systems that perform data science tasks through natural language interaction.

\begin{figure*}[t]  
    \centering
    \includegraphics[width=\textwidth]{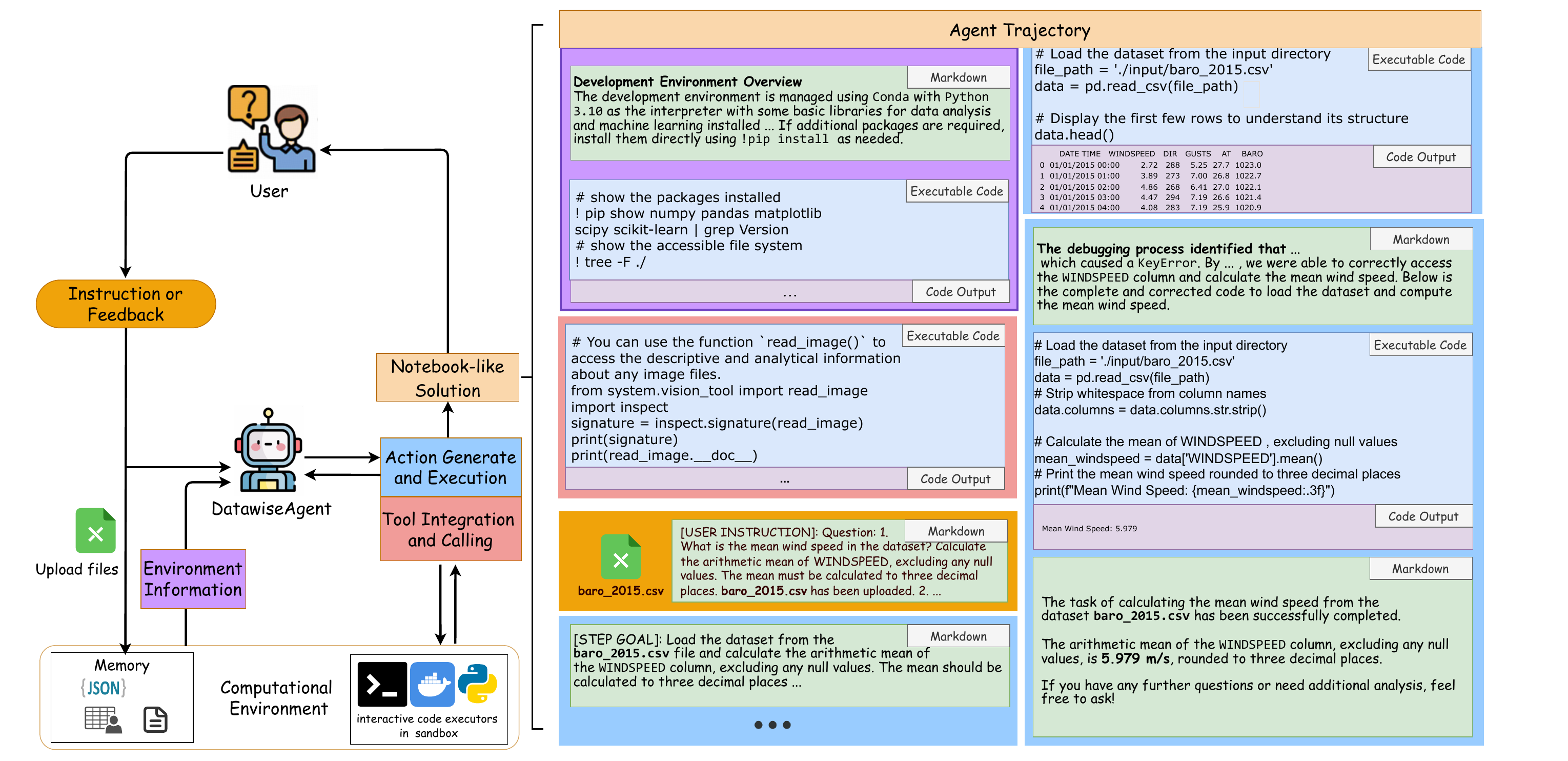} 
    \vspace{-1.8em}
    \caption{\textbf{\framework performs diverse data science tasks across various models by operating entirely within a computational notebook.} The unified interaction representation expresses all agent–user–environment communication. Tool integration involves importing external APIs or libraries via code cells, with tool descriptions provided in markdown; environment information, such as system details or resource status, is either proactively injected as markdown at initialization or obtained through code execution during task progress.} 
    \vspace{-1.5em}
    \label{fig:unified_interaction}  
\end{figure*}


However, current data science agents face three key limitations: 
(1) \textbf{Focus on Isolated Phases.} 
Many existing agents target specific stages of the data science pipeline, such as feature engineering~\cite{hollmann2024feature}, model selection~\cite{shen2024hugginggpt}, or hyperparameter tuning~\cite{zhang2023hyper} while overlooking the interdependent nature of real-world workflows. As a result, they fall short of supporting comprehensive end-to-end automation.
(2) \textbf{Limited Task and Model Adaptability.} Agents designed for broader workflows often struggle to generalize across diverse task types or model configurations~\cite{qiao2023taskweaver,hong2024data,hu2024infiagent}. 
While general-purpose frameworks such as ReAct~\cite{yao2023react, wang2024executable} and AutoGen~\cite{wu2023autogen} offer cross-domain applicability, they tend to exhibit suboptimal performance in specialized scenarios such as exploratory analysis or predictive modeling, particularly under constrained model capacities.
(3) \textbf{Over-Reliance on SOTA LLMs in Agent Design.} The majority of current data science agents are designed under the assumption of access to SOTA LLMs~\cite{hong2024data, qiao2023taskweaver}, such as GPT-4o. These systems often lack scalability and robustness when deployed with smaller or open-source models, limiting their applicability in resource-constrained or privacy-sensitive settings.

To address these limitations, we draw inspiration from the exploratory, progressive, iterative workflows that human data scientists follow in computational notebooks~\cite{head2019managing}. As the \emph{de facto} interface for data science, notebooks integrate natural language, code, and real-time feedback~\cite{rule2018exploration,chattopadhyay2020s, wang2022documentation,wang2021makes}. We posit that this paradigm provides a natural foundation for building adaptive and robust data science agents.

To this end, we propose \textbf{\framework}, a notebook-centric LLM agent framework designed for \textbf{adaptive} and \textbf{robust} data science automation (see Figure \ref{fig:unified_interaction}).
\framework combines two key components: (1) a \emph{unified interaction representation} that expresses all agent–user–environment communication as interleaved markdown and code cells within computational notebooks; and (2) a \emph{finite-state transducer (FST)-based multi-stage architecture} that governs agent behavior across four functional stages, including DFS-like planning, incremental execution, self-debugging, and post-filtering. 
This design enables flexible long-horizon planning, progressive solution development, and robust recovery from execution failures, making \framework suitable for deployment with LLMs of varying capacities and capabilities.

We evaluate \framework on three representative data science scenarios, namely data analysis, scientific visualization, and predictive modeling, across both proprietary (GPT-4o, GPT-4o mini) and open-source (Qwen2.5 at multiple scales) LLMs. 
Experimental results show that \framework consistently achieves SOTA performance under comparable evaluation conditions, surpassing strong baselines such as ReAct~\cite{yao2023react,hu2024infiagent}, MatplotAgent~\cite{yang-etal-2024-matplotagent}, AutoGen~\cite{wu2023autogen}, and Taskweaver~\cite{qiao2023taskweaver}.
Notably, on the challenging DSBench data modeling benchmark, \framework achieves over 90\% task success and more than 40 Relative Performance Gap (RPG) across all LLMs, including surpassing prior SOTA results even when using GPT-4o mini.
Further evaluation shows that \framework maintains strong performance on weaker LLMs and widens the performance gap with baseline methods, highlighting its robustness and scalability.   
In summary, these results demonstrate that \framework provides a practical and scalable foundation for robust, end-to-end data science automation across diverse tasks and LLM configurations.

\section{Related Work}
\label{sec:related_work}

\paragraph{LLMs for Code Generation.} 

Large Language Models (LLMs) have achieved strong performance across a range of code-related tasks~\cite{jiang2024codingsurvey}, including completion~\cite{li2023starcoder, roziere2023code}, translation~\cite{chen2021evaluating}, and repair~\cite{anthropic2024claude3, achiam2023gpt}. However, generating correct code in a single attempt remains challenging, particularly for complex or interactive tasks~\cite{chenteaching}. Recent studies show that external feedback and iterative refinement can significantly improve code generation~\cite{zhou2024language, zhong2024debug, madaan2024self, shinn2024reflexion}. Building on these findings, we focus on data science tasks and investigate how to leverage diverse LLMs' limited reasoning and coding capabilities, along with feedback, to enable adaptive end-to-end automation.

\paragraph{LLM-based Data Science Agents.}
LLM-based agents have shown promise in automating various stages of the data science pipeline, such as feature engineering~\cite{hollmann2024feature}, model selection~\cite{shen2024hugginggpt}, and hyperparameter tuning~\cite{zhang2023hyper}. To support broader workflows, a range of frameworks have been proposed for machine learning pipelines~\cite{guo2024dsagent, jiang2025aide, zhang2023automl, li2024autokaggle, trirat2024automl, zhang2024mlcopilot}, data analysis~\cite{qiao2023taskweaver, openai_advanced_data_analysis}, and visualization~\cite{yang-etal-2024-matplotagent}. While effective within specific scopes, many agents lack adaptability across tasks and models. In particular, most ML-focused agents~\cite{guo2024dsagent, jiang2025aide} adopt single-turn paradigms, limiting support for multi-turn interaction and human involvement. Furthermore, recent end-to-end systems~\cite{hong2024data} often rely on powerful proprietary LLMs, such as GPT-4o, limiting deployment with smaller or open-source models. In contrast, our agent framework supports robust, adaptive automation across diverse data science tasks and LLMs.
\section{\framework}

We present \framework, a novel notebook-centric LLM agent framework for effective, adaptive, and robust data science automation.
Inspired by how human data scientists work, through \emph{exploratory}, \emph{progressive}, and \emph{iterative} strategies within computational notebooks, \framework comprises two key components: (1) \emph{unified interaction representation} that captures all agent–user–environment communication via interleaved markdown and code cells; (2) \emph{finite-state transducer (FST)-based multi-stage architecture} that governs agent behavior via transitions across four core stages.
This architecture supports flexible planning, progressive solution development, and robust recovery from execution failures, making \framework well-suited for models with varying reasoning and coding capabilities.

\subsection{Unified Interaction Representation}
\label{subsec:unified_interaction_representation}

Computational notebooks are central to data science practice, seamlessly integrating natural language, code, and execution feedback. Inspired by this paradigm, \framework operates entirely within a notebook environment, enabling agents to reason, act, and revise solutions in a format familiar to practitioners.
To support this design, we define a \emph{unified} interaction representation (see Figure~\ref{fig:unified_interaction}) in which all agent–user–environment communication, including task instructions, environment information, tool integration and calling, and observations, is expressed as a sequence of \textbf{markdown} and \textbf{executable code cells}. Agents incrementally construct solutions by generating and updating cells over multiple rounds, producing an interpretable execution trace that supports user feedback and follow-up interaction.

Unlike prior systems that adopt stage-specific task formats (e.g., JSON-based graph planning or mixed-format tool calls)~\cite{qiao2023taskweaver, hong2024data, zhang2023data}, \framework maintains a structurally unified interaction mode, where context and actions are represented as cell sequences. We posit that this cell-level consistency reduces cognitive load and enhances in-context reasoning, particularly for models with constrained capabilities, while also enabling transparent oversight and unified multi-turn interaction.

\subsection{FST-Based Multi-Stage Architecture}
\label{subsec:fst-based_multi-stage_architecture}

\begin{figure}[t] 
    \centering
    \includegraphics[width=\linewidth]{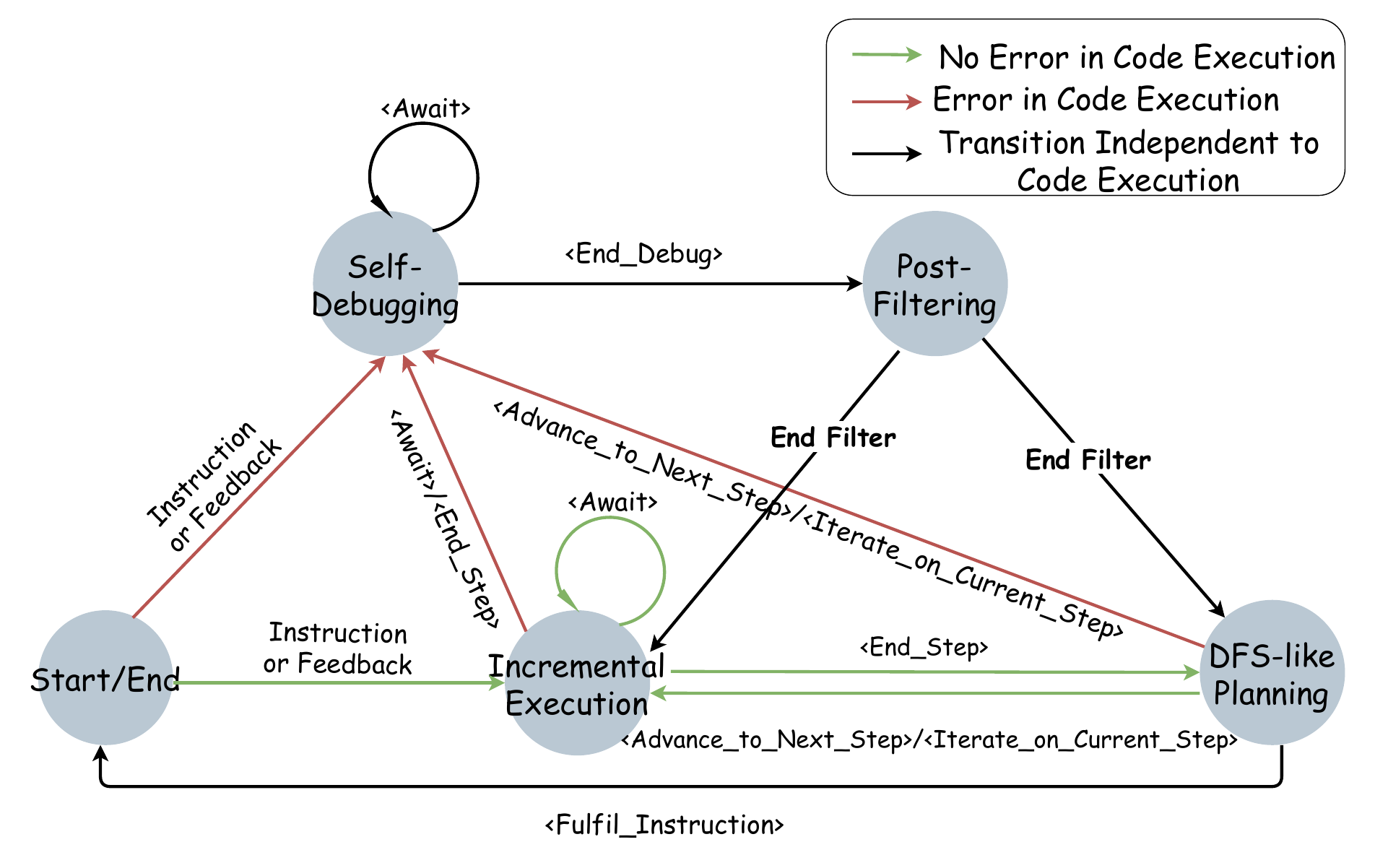} 
    \vspace{-2.5em}
    \caption{\textbf{State transition diagram of the FST-based multi-stage architecture}, modeled as a non-deterministic finite-state transducer (NFST). Transitions are driven by user instructions or feedback, agent-generated action signals, and execution feedback from the environment. At each state, the agent generates and executes actions based on the current context before proceeding to the next state.} 
    \vspace{-2.0em}
    \label{fig:state_transition} 
\end{figure}

To govern the agent’s behavior in a structured, modular, and extensible manner, \framework adopts a finite-state transducer (FST)-based multi-stage architecture. 
Rather than defining a rigid workflow or static pipeline, \framework organizes the problem-solving process into four core stages, including \emph{DFS-like planning}, \emph{incremental execution}, \emph{self-debugging}, and \emph{post-filtering} (detailed in~\cref{subsec:detail_stages}), and employs a finite-state transducer~\cite {hopcroft2001introduction, carroll1989theory} to orchestrate autonomous transitions among them.  Notably, this design facilitates modular extension with new stages and supports fine-grained ablation of individual components.

We conceptualize the agent as a non-deterministic finite-state transducer (NFST), where the state space $Q = \{q_{\text{plan}}, q_{\text{inc}}, q_{\text{debug}}, q_{\text{filter}}, q_0\}$ corresponds to the four functional stages and a special start/end state. The idle state $q_0$ denotes either task completion or readiness for new instructions. 
State transitions are driven by internally generated \emph{action signals} and external inputs, including user instructions and environment feedback (i.e., execution success or failure).
At each state, \framework produces two outputs: an \emph{action}, uniformly represented as markdown and executable code cells, and an \emph{action signal} indicating the intended next state. The action is executed in the notebook environment, yielding external feedback.
The agent determines its next state via the transition function $\delta(q, \sigma, f)$, which takes as input the current state $q$, the generated action signal $\sigma$, and the feedback $f$ from the environment or user.

\begin{algorithm}[b]
\caption{FST-based Multi-Stage Architecture}
\label{alg:workflow}
\begin{small}
\begin{algorithmic}[1]
\REQUIRE $I$: task input,\quad $\mathcal{H}$: context history, $\text{Agent}_\mathcal{P}$: LLM agent with language model $\mathcal{P}$

\STATE Initialize context: $\mathcal{H} \leftarrow$ environment info and tools
\STATE $\mathcal{H}.update(I)$,\quad $q \leftarrow q_0$,\quad $\sigma \leftarrow I$,\quad $f \leftarrow \texttt{no\_error}$

\WHILE{\text{True}}
    \STATE Generate action and action signal: A, $\texttt{signal} \leftarrow \text{Agent}_\mathcal{P}(q, \mathcal{H})$
    \STATE Execute action $A$ and receive feedback $f \in \{\texttt{error}, \texttt{no\_error}\}$
    \STATE Determine the next state: $q \leftarrow \delta(q, \sigma, f)$
    \STATE Update context with executed $A$: $\mathcal{H}.update(A, f)$
    \STATE Update action signal: $\sigma \leftarrow \texttt{signal}$

    \IF{$q = q_0$} 
        \STATE \textbf{Exit loop} (Task complete or waiting for new instructions)
    \ENDIF
    
\ENDWHILE
\RETURN $\mathcal{H}$
\end{algorithmic}
\end{small}
\end{algorithm}

The runtime logic of the FST-based architecture is formalized in~\cref{alg:workflow}. The agent starts in an idle state and, upon receiving a user instruction, autonomously transitions across functional stages, generating and executing actions, processing feedback, and updating context, until the task is completed. It then returns to the idle state, awaiting further user instructions or feedback. 
The state transition process is visualized in Figure~\ref{fig:state_transition} using an NFST formulation for clarity. The corresponding deterministic FST, stage-wise action signal spaces, and implementation details are provided in~\cref{sec:appendix-fst}.



\subsection{Detailed Explanation of Each Stage}
\label{subsec:detail_stages}
To operationalize FST-based multi-stage architecture, \framework organizes the agent’s behavior into four functional stages: DFS-like planning, incremental execution, self-debugging, and post-filtering. These stages are inspired by how data scientists work in notebooks, collectively supporting flexible planning, progressive solution development, and robust recovery from execution failures.

\begin{figure*}[htbp]  
    \centering
    \includegraphics[width=0.90\textwidth]{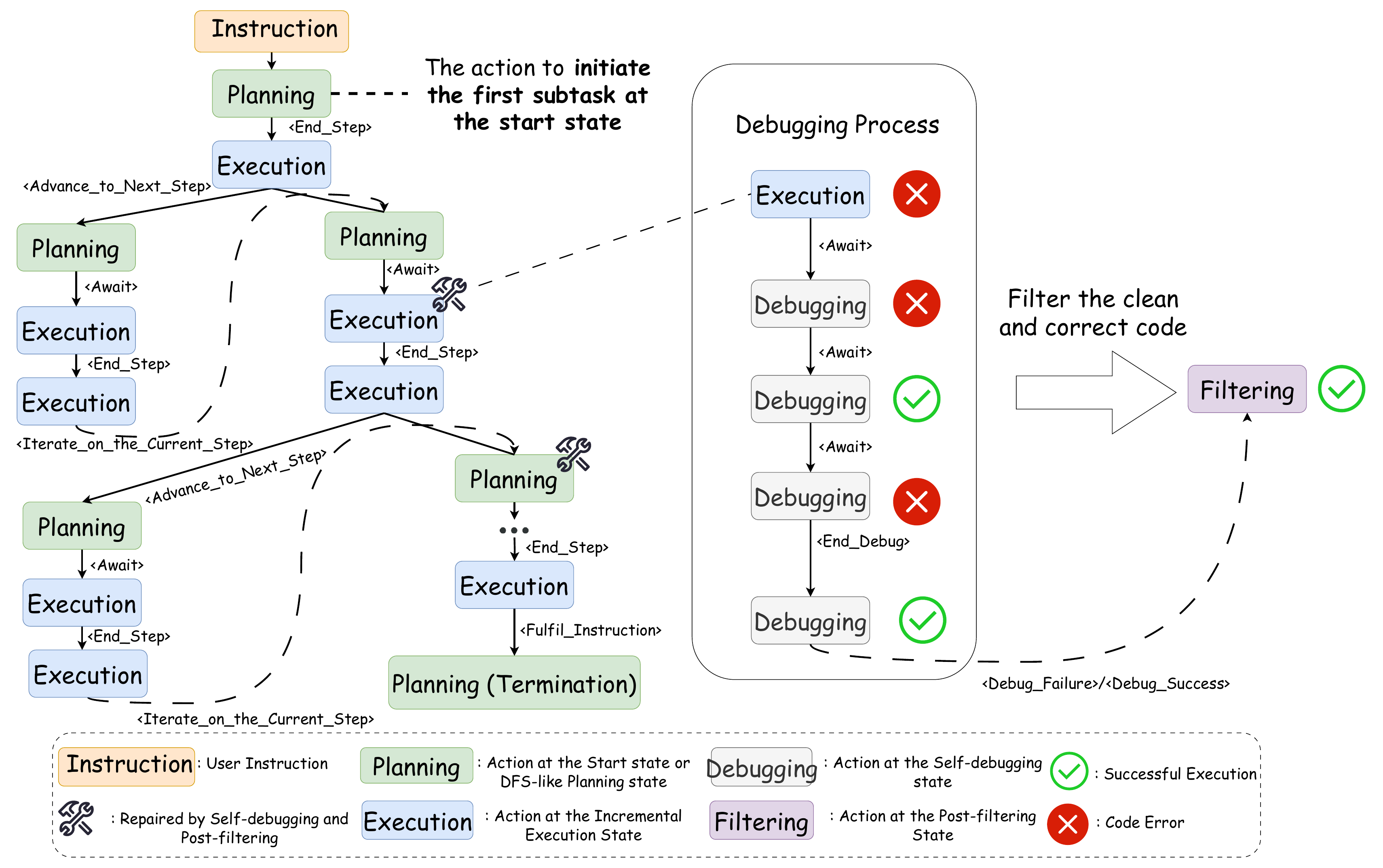} 
    \vspace{-1.0em}
    \caption{\textbf{Illustration of \framework’s task-completion process.} Left: tree-structured trajectory from DFS-like planning and incremental execution. Right: code repair via self-debugging and post-filtering.} 
    \vspace{-1.0em}
    \label{fig:planning-debugging}  
\end{figure*}
\paragraph{DFS-like Planning and Incremental Execution.} 
\framework introduces two tightly coupled stages: DFS-like planning and incremental execution. Together, they form a tree-structured task-completion process (see Figure~\ref{fig:planning-debugging}), enabling flexible exploration and progressive problem solving under constrained reasoning and coding capabilities.

In the DFS-like planning stage, the agent dynamically selects one of three actions based on task progress and feedback: (1) \emph{advance} to the next subgoal; (2) \emph{backtrack} to revise the current subtask by replacing it with a newly proposed one; or (3) \emph{terminate} when the objective is satisfied. 
This non-linear planning strategy departs from static or sequential pipelines, enabling adaptive exploration of alternative solution paths.
During incremental execution, instead of one-shot generation followed by iterative refinement, each subtask is completed step by step through interleaved markdown and code cells, leveraging fine-grained feedback. This progressive strategy exploits limited model capabilities while improving robustness against execution failures.

By coordinating planning and execution via transitions between $q_\text{plan}$ and $q_\text{inc}$, \framework enables models of varying capabilities to perform long-horizon reasoning and adaptively solve complex data science tasks through progressive strategies.

\paragraph{Code Repair through Self-Debugging and Post-Filtering.} 
To ensure robust recovery from execution failures and prevent the accumulation of misleading traces in the context, \framework introduces a code repair module implemented via FST transitions across two stages: self-debugging and post-filtering (see Figure~\ref{fig:planning-debugging}).

In the self-debugging stage, the agent analyzes and iteratively refines faulty code using execution feedback. This stage is designed to be extensible, allowing integration of advanced LLM-based repair techniques~\cite{hu2024leveraging, chenteaching, zhong2024debug} to further enhance correction performance. The post-filtering stage then assesses whether the error has been resolved in the debugging process: if successful, the agent extracts the clean and corrected code from the debugging trace; otherwise, it generates a concise diagnostic report in markdown, distilling key failure insights to prevent context pollution and guide future decisions.

This code repair module is triggered by execution errors during DFS-like planning or incremental execution. Upon completion, the agent replaces the original faulty code and debugging traces with post-filtered output, and resumes the ongoing task-solving process.
\section{Experiments}
\label{sec:experiments}


In this section, We evaluate \framework across three key dimensions: its \emph{effectiveness} and \emph{adaptability} across tasks and LLMs (\cref{subsec: adaptability}), its \emph{robustness} under varying model capabilities and scales (\cref{subsec:robustness}), and the contributions of its planning and code repair modules through an ablation study (\cref{ablation}).

\subsection{Experimental Setup}

\paragraph{Benchmarks, Evaluation Metrics, and Baselines.} We evaluate \framework on three public benchmarks covering core data science scenarios, each with tailored metrics and established baselines: 

(1) \textbf{Data Analysis:} InfiAgent-DABench~\cite{hu2024infiagent} contains 257 challenges with CSV inputs and multi-level (easy/medium/hard) analysis questions. We report \emph{Accuracy by Questions} (ABQ), i.e., the proportion of correctly answered questions. Baselines include ReAct~\cite{yao2023react, hu2024infiagent}, AutoGen~\cite{wu2023autogen}, TaskWeaver~\cite{qiao2023taskweaver}, and Data Interpreter~\cite{hong2024data}.

(2) \textbf{Scientific Visualization:} We use MatplotBench~\cite{yang-etal-2024-matplotagent}, comprising 100 expert-verified cases involving input data, user queries, and reference plots. A vision model assigns a 0–100 score based on alignment with ground truth. We use GPT-4o as a unified scoring model across all settings. 
Baselines include Direct Decoding, where the model generates code in a single pass; MatplotAgent~\cite{yang-etal-2024-matplotagent}, a vision-augmented agent specialized in plotting tasks; and AutoGen~\cite{wu2023autogen}. 

(3) \textbf{Predictive Modeling:} We use the data modeling part from DSBench~\cite{jing2024dsbench}, which includes 74 real-world Kaggle competitions. Each task requires predictive modeling based on training/testing data, a sample submission file, and a detailed description.
Following DSBench~\cite{jing2024dsbench}, we report \emph{Task Success Rate}, \emph{Relative Performance Gap} (RPG).
RPG serves as a standardized score that reflects the agent’s overall performance across tasks by directly evaluating the performance of the resulting models on testing datasets. A task is marked incomplete if it exceeds the 3600-second time limit.
We compare \framework with results reported for AutoGen \cite{wu2023autogen} and Code Interpreter\footnote{\url{https://platform.openai.com/docs/assistants/tools/code-interpreter}}.

Further details on the benchmarks, metric definitions, and method configurations are provided in~\cref{sec:appendix-experiment-details}.

\begin{table}[t]
    \centering
    \small
    \begin{tabular}{l|l|c}
        \toprule
        \textbf{Model} & \textbf{Method} & \textbf{ABQ/\% $\uparrow$} \\
        \midrule
        \multirow{5}{*}{GPT-4o mini}
        & ReAct & \underline{80.08} \\
        & AutoGen & 70.04 \\
        & Taskweaver & 76.65 \\
        & Data Interpreter & 67.70 \\
        & \textbf{\framework (Ours)} & \textbf{82.88} \\
        \midrule
        \multirow{6}{*}{GPT-4o}
        & ReAct & \underline{81.32} \\
        & AutoGen & 73.54 \\
        & Taskweaver & \textbf{85.99} \\
        & \textcolor{gray}{Data Interpreter}$^{*}$ & \textcolor{gray}{94.93$^{*}$}\\
        & Data Interpreter & 75.78 \\
        & \textbf{\framework (Ours)} & \textbf{85.99} \\
        \midrule
        \multirow{3}{*}{\makecell[l]{Qwen2.5-72B\\-Instruct}}
        & ReAct & \underline{75.88} \\
        & AutoGen & 70.04 \\
        & Taskweaver & 74.71 \\
        & \textbf{\framework (Ours)} & \textbf{81.71} \\
        \bottomrule
    \end{tabular}
    \vspace{-0.7em}
    \caption{\textbf{Performance comparison on InfiAgent-DABench.} Asterisked ($^{*}$) result is from \citet{hong2024data} and could not be reproduced in our setting; it is shown for reference only and excluded from SOTA comparison. Best results in bold; second-best underlined.}
    \vspace{-1.8em}
    \label{tab:infiagent}
\end{table}

\begin{table}[t]
    \renewcommand{\arraystretch}{0.95}
    \small
    \centering
    \footnotesize
    \setlength{\tabcolsep}{3.5pt}
    \begin{adjustbox}{max width=\linewidth}
    \begin{tabular}{llcc}
        \toprule
        \textbf{Model} & \textbf{Framework} & \textbf{Avg. Score $\uparrow$} & \textbf{$\Delta$ Score $\uparrow$} \\
        \midrule
        \multirow{6}{*}{\makecell[l]{GPT-4o\\mini}} 
        & Direct Decoding & 38.09 & - \\
        & MatplotAgent & 51.44 & +13.35 \\
        & AutoGen & 51.82 & +13.73 \\
        & \quad w/ visual tool & 52.07 & +13.98 \\
        & \textbf{\framework} & \underline{55.85} & \underline{+17.76} \\
        & \textbf{\quad w/ visual tool} & \textbf{58.60} & \textbf{+20.51} \\
        \midrule
        \multirow{6}{*}{GPT-4o} 
        & Direct Decoding & 45.28 & - \\
        & MatplotAgent & 57.86 & +12.58 \\
        & AutoGen & 60.42 & +15.14 \\
        & \quad w/ visual tool & \underline{63.60} & \underline{+18.32} \\
        & \textbf{\framework} & 61.22 & +15.94 \\
        & \textbf{\quad w/ visual tool} & \textbf{64.33} & \textbf{+19.05} \\
        \midrule
        \multirow{5}{*}{\makecell[l]{Qwen2.5\\-72B-Instruct}} 
        & Direct Decoding & 47.54 & - \\
        & AutoGen & 40.80 & -6.74 \\
        & \quad w/ visual tool & 53.72 & +6.18 \\
        & \textbf{\framework} & \underline{56.41} & \underline{+8.87} \\
        & \textbf{\quad w/ visual tool} & \textbf{61.88} & \textbf{+14.34} \\
        \bottomrule
    \end{tabular}
    \end{adjustbox}
    \vspace{-0.5em}
    \caption{\textbf{Performance comparison on MatplotBench.} \textbf{$\Delta$ Score} denotes the score gain over Direct Decoding. Bold and underline highlight the best and second-best results, respectively. Visual tool rows indicate integration with the GPT-4o mini-based visual tool.}
    \vspace{-0.8em}
    \label{tab:matplotbench}
\end{table}

\begin{table}[t]
    \renewcommand{\arraystretch}{0.95}
    \small
    \centering
    \footnotesize
    \setlength{\tabcolsep}{3.5pt}
    \begin{adjustbox}{max width=\linewidth}
    \begin{tabular}{llcc}
        \toprule
        \textbf{Framework} & \textbf{Model} & \textbf{Task Success/\%} & \textbf{RPG} \\
        \midrule
        \multirow{5}{*}{AutoGen} 
        & Llama3-8B & 5.41 & 1.55 \\
        & Llama3-70B & 16.22 & 7.79 \\
        & GPT-4 & 87.84 & 45.52 \\
        & GPT-4o & 71.62 & 34.74 \\
        & GPT-4o mini & 22.97 & 11.24 \\
        \midrule
        \multirow{3}{*}{Code Interpreter} 
        & GPT-4 & 54.05 & 26.14 \\
        & GPT-4o & 44.59 & 19.87 \\
        & GPT-4o mini & 39.19 & 16.90 \\
        \midrule
        Human$^*$ & Human$^*$ & 100.00$^*$ & 65.02$^*$ \\
        \midrule
        \multirow{3}{*}{\textbf{\framework}} 
        & GPT-4o & \textbf{98.64} & \textbf{53.18} \\
        & GPT-4o mini & \textbf{98.64} & \underline{46.61} \\
        & Qwen2.5-72B & \underline{91.89} & 42.90 \\
        \bottomrule
    \end{tabular}
    \end{adjustbox}
    \vspace{-0.5em}
    \caption{\textbf{Performance comparison on 74 Data Modeling tasks from DSBench.} Bold and underline indicate best and second-best results. *Human performance is from \cite{jing2024dsbench}, based on evaluations across 22 competitions.}
    \vspace{-1.5em}
    \label{tab:datamodeling}
\end{table}

\paragraph{Model Configurations.}
To assess the adaptability and generalization across diverse LLMs, we evaluate \framework using both proprietary and open-source models: GPT-4o, GPT-4o mini~\cite{hurst2024gpt}, and Qwen2.5-72B-Instruct~\cite{yang2024qwen2} \footnote{\texttt{gpt-4o-2024-08-06} and \texttt{gpt-4o-mini-2024-07-18}.}. 
To further examine robustness under varying model capacities, we conduct additional experiments on InfiAgent-DABench~\cite{hu2024infiagent} using Qwen2.5 instruction-tuned models of different sizes: 7B, 14B, 32B, and 72B. 


\subsection{Effectiveness and Adaptability Across Tasks and LLMs}
\label{subsec: adaptability}
We evaluate \framework's effectiveness and adaptability across three representative data science scenarios: data analysis, scientific visualization, and predictive modeling, using three distinct LLMs: GPT-4o, GPT-4o mini, and Qwen2.5-72B-Instruct.
The comparative performance with existing agent frameworks is presented in Tables~\ref{tab:infiagent}, \ref{tab:matplotbench}, and~\ref{tab:datamodeling}.

\begin{figure}[t]
    \centering
    \includegraphics[width=\linewidth]{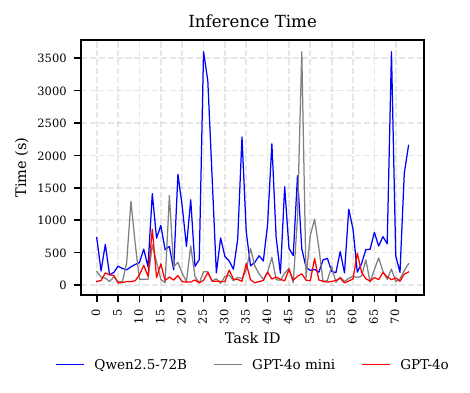}
    \small
    \vspace{-2.7em}
    \caption{Inference time of \framework on 74 data modeling tasks from DSBench.}
    \vspace{-1.5em}
    \label{fig:inference_time}
\end{figure}
\begin{figure}[tbp]
    \centering
    \includegraphics[width=\linewidth]{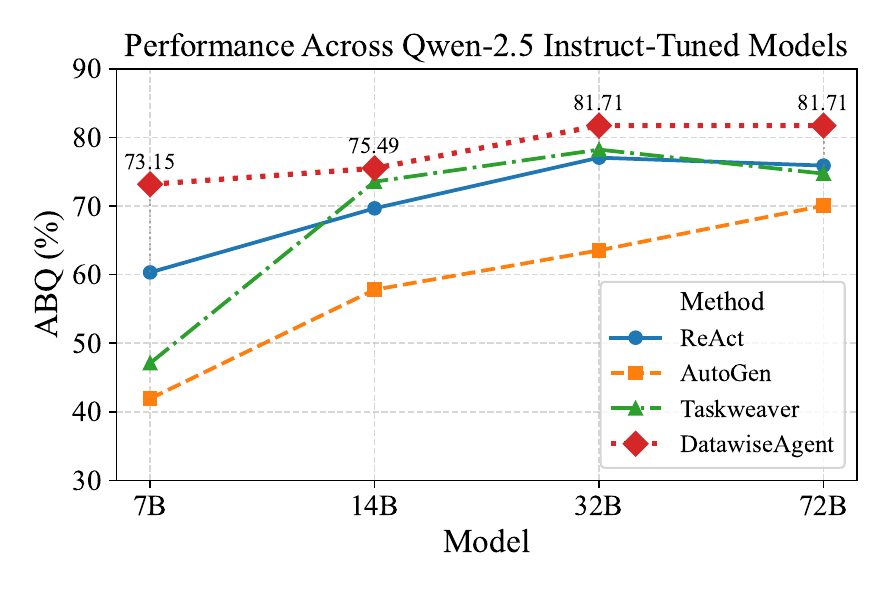}
    \small
    \vspace{-3.0em}
    \caption{\textbf{Performance across Qwen2.5 models on InfiAgent-DABench.} \framework demonstrates strong robustness across models of different sizes, maintaining top performance while the gap over competing methods becomes more pronounced on smaller models.} 
    \vspace{-2.0em}
    \label{fig:agent_robustness}  
\end{figure}

\paragraph{Data Analysis.} 
In data analysis, as shown in ~\cref{tab:infiagent}, \framework achieves strong performance across all model settings, highlighting its strong capability in executing accurate and reliable data analyses. On GPT-4o mini and Qwen2.5-72B-Instruct, \framework outperforms all baselines, achieving SOTA results. On GPT-4o,  \framework matches Taskweaver, a framework specifically designed for data analysis tasks, while surpassing AutoGen and ReAct. Notably, although Data Interpreter is reported to reach 94.93\% on GPT-4o by \citet{hong2024data}, our best-effort replication under comparable conditions yields significantly lower scores (75.78\% on GPT-4o and 67.7\% on GPT-4o mini), which we include for fair comparison. This discrepancy may be due to differences in evaluation settings not fully specified in the original paper. We include both results for transparency.

\paragraph{Scientific Visualization.} 
In scientific visualization, \framework consistently achieves the best performance across models (as shown in ~\cref{tab:matplotbench}), highlighting its ability to produce high-quality scientific visual output. On GPT-4o, \framework obtains the highest average score of 64.33, both with and without the visual tool (61.22 without, 64.33 with), establishing a new SOTA. 
We also observe that \framework leads by a clear margin in completion rate and high-quality output proportion across all settings, demonstrating strong robustness in producing valid and reliable figures. These auxiliary metrics are reported in Appendix Table~\ref{tab:full_matplotbench}.

To further assess tool usage impact, we incorporate a GPT-4o mini-based visual tool into AutoGen and \framework, enabling iterative figure refinement via visual-textual feedback. Implementation details are provided in~\cref{subsec:appendix-visual_tool}. Tool-integrated variants consistently outperform their non-tool counterparts, aligned with findings from ~\cite{yang-etal-2024-matplotagent}. Notably, \framework with visual tool integration achieves the best results in all three model configurations, suggesting the effectiveness and sound design of our tool integration.

\begin{table}[htbp]
\centering
\scriptsize
\begin{tabular}{l|c|c|c}
\toprule
Model & Avg. LLM Calls & Planning & Code Repair \\
\midrule
GPT-4o & 12.31 & 4.72 & 0.62 \\
GPT-4o-mini & 18.80 & 4.15 & 1.39 \\
Qwen2.5-72B-Instruct & 16.91 & 5.05 & 1.18 \\
\bottomrule
\end{tabular}
\vspace{-0.7em}
\small
\caption{\textbf{Average number of transitions per task in \framework on 74 Data Modeling tasks from DSBench.} “Avg. LLM Calls” counts the number of LLM calls, while “Planning” and “Code Repair” refer to transitions into the respective modules.}
\vspace{-1.5em}
\label{tab:transition_statistics}
\end{table}

\begin{table}[htbp]
\centering
\small
\setlength{\tabcolsep}{3pt}
\begin{adjustbox}{max width=\linewidth}
\begin{tabular}{l|ccc|c}
\toprule
\textbf{Method} & \textbf{Inf. /s} & \textbf{Suc. /\% $\uparrow$} & \textbf{RPG $\uparrow$} & \textbf{ABQ /\% $\uparrow$} \\
\midrule
\framework & 291.57 & \textbf{98.64} & \textbf{46.61} & \textbf{77.14} \\
\quad w/o planning & 529.99 & 77.03 & 38.35 & 70.86 \\
\quad w/o code repair & 429.18 & 87.84 & 43.80 & 75.43 \\
\bottomrule
\end{tabular}
\end{adjustbox}
\vspace{-0.5em}
\caption{\textbf{Ablation results of \framework on GPT-4o mini.} Metrics are reported for 74 data modeling tasks from DSBench and 175 medium-/hard-level data analysis tasks from InfiAgent-DABench. 
\textbf{Inf. /s} = Inference time, which measures the average time taken to complete a single task; 
\textbf{Suc. /$\%$} = Task Success rate.}
\vspace{-1.5em}
\label{tab:ablation}
\end{table}

\paragraph{Predictive Modeling.} 
In predictive modeling, \framework achieves SOTA performance across all model settings, as shown in ~\cref{tab:datamodeling}, demonstrating strong capability in solving comprehensive and complex end-to-end data-centric prediction tasks. 
It consistently obtains high Task Success Rates ($\geq$90\%) and strong RPG values, with GPT-4o reaching the best overall performance (RPG 53.18). 
We further observe that \framework with the weaker GPT-4o mini outperforms AutoGen with GPT-4, suggesting the potential for achieving competitive performance with smaller models.

Together, the results demonstrate the strong performance and adaptability of \framework across diverse tasks and LLMs. It achieves SOTA results in both scientific visualization and predictive modeling, while maintaining strong performance in data analysis. Additionally, these results reinforce the effectiveness of \framework's tool integration design, particularly in scientific visualization.

We also observe high task completion rates across domains (see~\cref{tab:datamodeling}; Appendix Table~\ref{tab:full_matplotbench}), which we attribute to \framework’s FST-based multi-stage architecture orchestrating DFS-like planning, incremental execution, and code repair. This design supports flexible long-horizon planning, progressive solution building, and robust recovery from failures.

\subsection{Robustness to Model Capability and Scale Variations}
\label{subsec:robustness}
\paragraph{Robustness to Model Capability.} Despite substantial differences in model capability, \framework achieves comparable performance across GPT-4o, GPT-4o mini, and Qwen2.5-72B-Instruct on predictive modeling tasks.
To understand this robustness, we analyze per-task inference time (i.e., time to complete a task) across models (Figure~\ref{fig:inference_time}), finding that Qwen2.5-72B-Instruct incurs significantly longer durations despite strong performance. Manual inspection reveals that this discrepancy primarily stems from inefficient code execution—especially in data preprocessing and model training. This suggests that stronger models like GPT-4o tend to generate more efficient code, leading to faster execution. To explain how weaker models nonetheless maintain performance, we examine \framework's internal state transitions (\cref{tab:transition_statistics}), including average LLM calls, planning steps, and code repair attempts per task. We find that weaker models invoke these modules more frequently, indicating that \framework dynamically adaptively increases reasoning depth and self-correction to compensate for limited model capability. These results highlight \framework’s robustness in adapting to varying model capabilities while maintaining competitive performance.

\paragraph{Robustness to Model Scale.} We further test robustness by evaluating \framework on Qwen2.5 instruct-tuned models of varying sizes (7B, 14B, 32B, 72B) using InfiAgent-DABench. As shown in Figure~\ref{fig:agent_robustness}, although all agent frameworks degrade with smaller models, \framework consistently outperforms all baselines on all scales. Notably, the performance gap between \framework and other methods \textbf{widens substantially} as model size decreases, demonstrating \framework’s robustness to model scale variation and its superior scalability compared to existing frameworks.

\subsection{Ablation Study of Planning and Code Repair Modules}
\label{ablation}
One key contribution of \framework is its FST-based multi-stage architecture. To assess the impact of key components, we ablate two modules, DFS-like planning and code repair (self-debugging and post-filtering), on 175 medium- and hard-level cases from InfiAgent-DABench and the Data Modeling tasks in DSBench, using GPT-4o mini. We compare three variants: (1) \textbf{\framework}: full system; (2) \textbf{w/o planning}: removes DFS-like planning, enforcing linear execution; (3) \textbf{w/o code repair}: disables code repair by removing self-debugging and post-filtering. 

As shown in \cref{tab:ablation}, performance consistently declines when either module is removed, with a more pronounced drop in predictive modeling, a more challenging task. These results underscore the importance of both flexible planning and robust recovery from failures in enabling \framework to solve complex data science tasks. Due to space limitations, we defer cost analysis and case study of \framework to ~\cref{sec:appendix-cost_analysis} and ~\cref{sec:appendix-case-study}.

\section{Conclusion}
We propose \framework, a notebook-centric LLM agent framework for adaptive and robust data science automation. By combining a unified interaction representation with an FST-based multi-stage architecture, \framework supports flexible long-horizon planning, progressive solution development, and robust recovery from execution failures within computational notebooks.
Experiments across diverse tasks and LLMs demonstrate its strong performance, adaptability, robustness across domains and models, establishing a notebook-centric paradigm for adaptive and robust data science automation.


\section{Limitations}
While \framework demonstrates effectiveness, adaptability, and robustness across multiple tasks and LLMs, several limitations remain that suggest directions for future work.
First, our evaluation of tool integration is limited to a single visual feedback tool in scientific visualization; broader assessment in domains with proprietary or complex toolchains (e.g., healthcare or finance) is needed. 
Second, although \framework is naturally suited for integration into computational notebooks (e.g., Jupyter or Colab), we do not evaluate human-in-the-loop collaboration. This omission reflects our focus on autonomous task completion, but evaluating interactive workflows remains a valuable and methodologically challenging direction for future work. 
These limitations point to promising directions for expanding \framework toward broader applicability in real-world and collaborative data science workflows.



\section{Ethics Statement}
This work does not involve human subjects, personal data, or proprietary user information. All datasets used in our experiments are publicly available. As \framework is designed to automate data science workflows through LLM-based agents, we acknowledge potential risks related to error propagation, unintended behavior, or unsafe code execution in autonomous settings. We encourage responsible and lawful use of such systems and recommend incorporating appropriate safeguards before real-world deployment. Our methodology is fully transparent and reproducible, and we support continued dialogue around fairness, accountability, and reliability in LLM-based automation.

\section{Acknowledgement}
This paper was supported by National Key R\&D Program of China
(No. 2023YFC3502902, 2021YFF1201100), National Natural Science
Foundation of China under Grants (62436006), Sanya Science and Technology Special Fund (No. 2024KFJX04) and Beijing Natural Science Foundation (No. L257018, No. L246024).


\bibliography{custom}
\clearpage

\appendix

\section{Cost Analysis}
\label{sec:appendix-cost_analysis}
\begin{table}[htbp]
\centering
\small
\begin{tabular}{lcccc}
\toprule
\textbf{Model} & \textbf{Framework} & \textbf{Cost/\$} & \textbf{ABQ/\% $\uparrow$} \\
\midrule
\multirow{4}{*}{GPT-4o} 
    & ReAct       & 4.79 & 81.32 \\
    & AutoGen     & 4.71 & 73.54 \\
    & TaskWeaver  & 16.19 & 85.99 \\
    & \textbf{\framework}  & 10.60 & 85.99 \\
\midrule
\multirow{4}{*}{GPT-4o mini} 
    & ReAct       & 0.26 & 80.08 \\
    & AutoGen     & 0.44 & 70.04 \\
    & TaskWeaver  & 1.64 & 76.65 \\
    & \textbf{\framework}  & 1.14 & 82.88 \\
\bottomrule
\end{tabular}
\caption{Total cost comparison on GPT-4o and GPT-4o mini on InfiAgent-DABench across 257 cases.}
\label{tab:cost_comparison_infiagent}
\end{table}
\begin{table*}[htbp]
    \renewcommand{\arraystretch}{0.86}  
    \small
    \centering
    \resizebox{\linewidth}{!}{
    \begin{tabular}{ll|cccc}
        \toprule
        \textbf{Framework} & \textbf{Model} & \textbf{Cost/\$}  & \textbf{Inference Time/s} & \textbf{Task Success/\% $\uparrow$} & \textbf{RPG $\uparrow$} \\
        \midrule
        \multirow{5}{*}{AutoGen} & Llama3-8b & - &  50.9 & 5.41 & 1.55 \\
        & Llama3-70b & - &  158.4 & 16.22 & 7.79 \\
        & GPT-4 & 19.34 &  77.4 & 87.84 & 45.52 \\
        & GPT-4o & 12.27 & 104.1 & 71.62 & 34.74 \\
        & GPT-4o mini &  0.10 & 26.7 & 22.97 & 11.24 \\
        \midrule
        \multirow{3}{*}{Code Interpreter} & GPT-4 &  38.81 &  237.6 & 54.05 & 26.14 \\
        & GPT-4o & 19.26 & 268.6 & 44.59 & 19.87 \\
        & GPT-4o mini &  2.70 & 199.6 & 39.19 & 16.90 \\
        \midrule
        \multirow{1}{*}{Human$^{*}$} & Human$^{*}$ & -& - & 100.00 & 65.02 \\
        \midrule
        \multirow{3}{*}{\textbf{\framework(Ours)}}& GPT-4o & 18.49 & 123.86 & \textbf{98.64} & \textbf{53.18} \\
        & GPT-4o mini & 2.13 & 291.57 & \textbf{98.64} & \underline{46.61} \\
        & Qwen2.5-72B-Instruct & - & 760.25  & \underline{91.89} & 42.9 \\
        \bottomrule
    \end{tabular}
    }
    \vspace{-0.7em}
    \caption{\textbf{Performance comparison on 74 Data Modeling tasks from DSBench.} We report \textbf{Cost}, \textbf{Inference Time}, \textbf{Task Success}, and \textbf{Relative Performance Gap (RPG)}, with best and second-best results in bold and underline, respectively. *Human performance is from \cite{jing2024dsbench}, based on evaluations across 22 competitions.}
    \label{tab:full_datamodeling}
\end{table*}
We record the total cost of \framework across different model settings, as illustrated in ~\cref{tab:full_datamodeling}. Compared to the previous best method, AutoGen with GPT-4, which incurs a cost of \$19.34, \framework outperforms it with a cost of only \$2.13, achieving superior performance. 
Additionally, when \framework with GPT-4o achieves the best performance of 53.18 in RPG, it incurs a cost of \$18.49. 
This demonstrates that \framework achieves strong performance more cost-effectively, delivering impressive results without incurring the high costs associated with previous approaches.

In addition to DSBench, we further evaluate the total cost on InfiAgent-DABench, which consists of 257 decision-making tasks. As shown in~\cref{tab:cost_comparison_infiagent}, \framework consistently demonstrates high cost-efficiency across different model settings.

Under the GPT-4o configuration, \framework achieves an ABQ score of 85.99\%, matching the best-performing baseline, but at a significantly lower cost (\$10.60 vs. \$16.19). Similarly, under the GPT-4o mini setting, \framework achieves the highest ABQ score of 82.88\%, while incurring only \$1.14 in total cost—substantially cheaper than TaskWeaver (\$1.64) and considerably more accurate than AutoGen.

These results highlight that \framework not only performs competitively or better in terms of quality, but also offers a favorable cost-performance trade-off, especially in scenarios requiring high scalability or low-latency inference.
\begin{figure*}[t] 
    \centering
    \includegraphics[width=\textwidth]{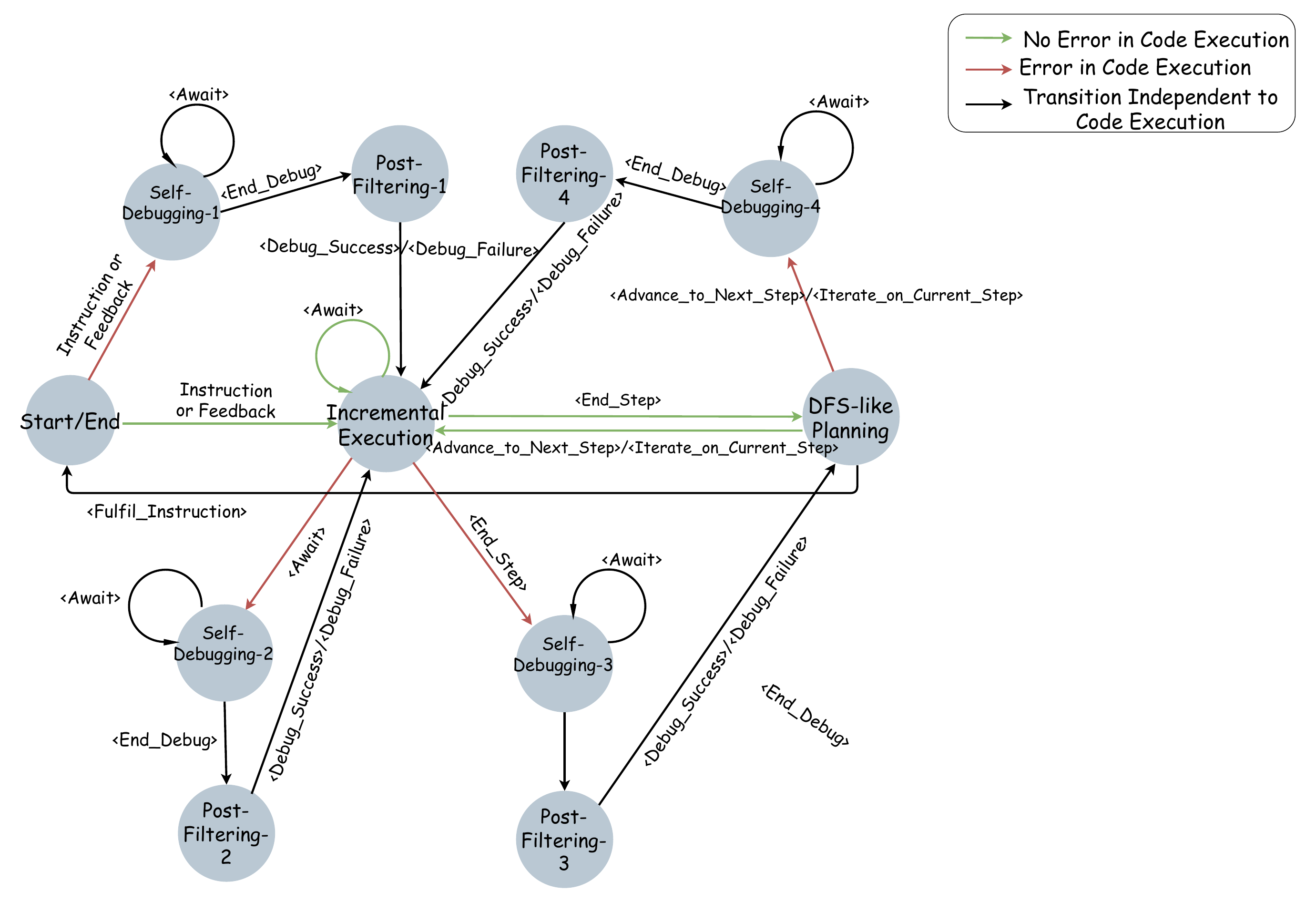} 
    \caption{State transition diagram in the FST-based multi-stage design of \framework, represented as a Non-deterministic Finite State Transducer (NFST). State transitions are driven by instructions or user feedback, action signals from the agent, and code execution feedback from the environment. Before each state transition, the agent generates and executes actions based on the current state.} 
    \label{fig:appendix_state_transition} 
\end{figure*}
\begin{table*}[htbp]
\centering
\small
\begin{tabular}{l|l}
\toprule
\textbf{Stage} & \textbf{Action Signal Space} \\
\midrule
DFS-like Planning & $\{\text{<Advance\_to\_Next\_Step>}, \text{<Iterate\_on\_the\_Current\_Step>},\text{<Fulfil\_Instruction>}\}$ \\
Incremental Execution & $\{\text{<Await>}, \text{<End\_Step>}\}$\\
Self-debugging & $\{\text{<Await>}, \text{<End\_Debug>}\}$ \\
Post-filtering & $\{\text{<Debug\_Failure>}, \text{<Debug\_Success>}\}$ \\
\bottomrule
\end{tabular}
\caption{
Action signal space for each stage of the \framework framework. At every stage, \framework generates an action and selects a corresponding signal from the defined signal space.}
\label{tab:appendix-state_action_signal}
\end{table*}

\section{Case Study on Predictive Modeling}
\label{sec:appendix-case-study}
\begin{figure*}[htbp]  
    \centering
    \includegraphics[width=\textwidth]{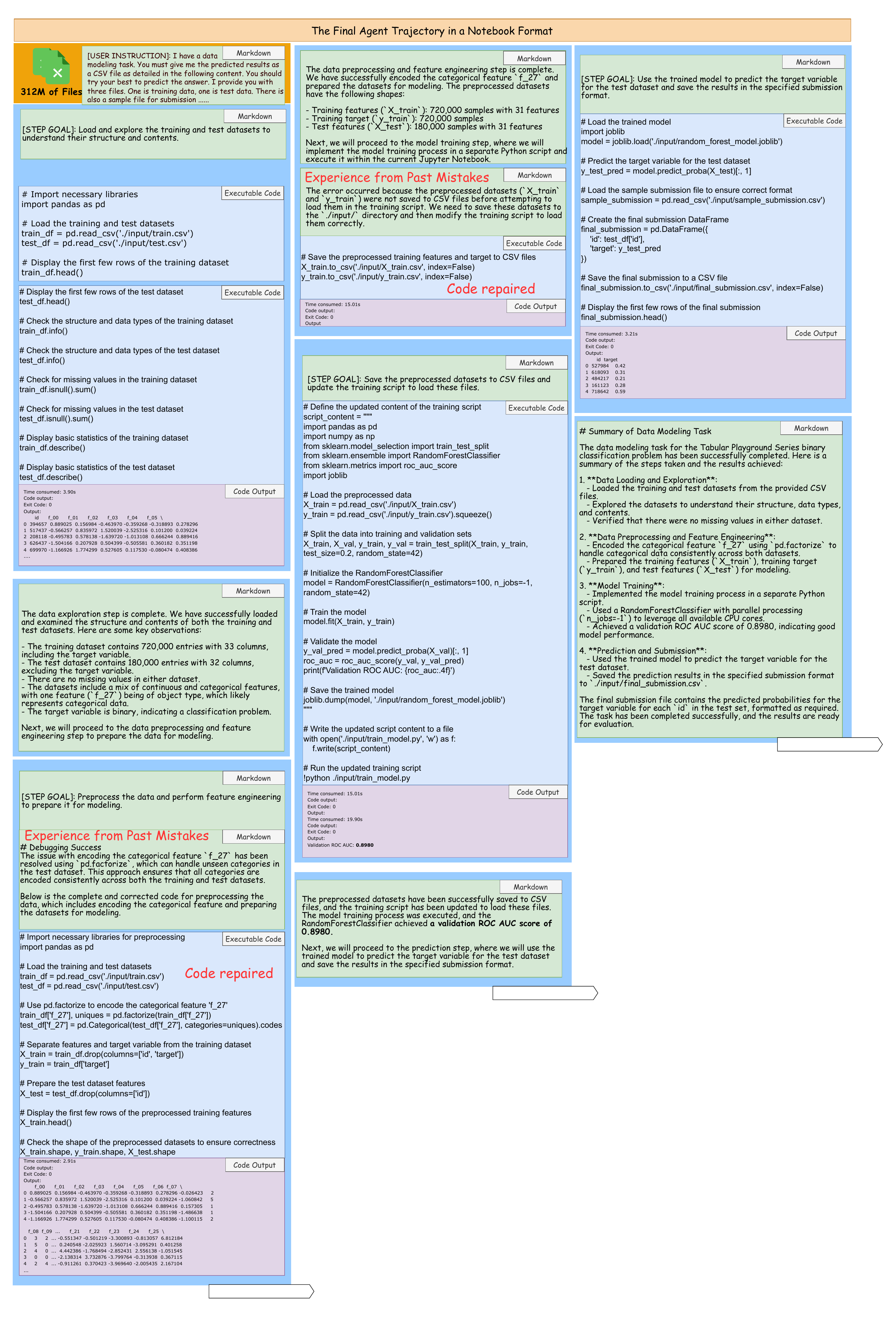} 
    \caption{The case example of \framework for the data modeling task with index = 48} 
    \label{fig:data_modeling_case_example}  
\end{figure*}

We present a case example corresponding to the data modeling task with index 48 from DSBench~\cite{jing2024dsbench}. 
The task’s instruction is shown in Figure~\ref{fig:instruction_case_example_data_modeling}, and the final agent trajectory of \framework with GPT-4o is illustrated in Figure~\ref{fig:data_modeling_case_example}. 
As demonstrated by this example, \framework utilizes DFS-like planning and incremental execution to dynamically decompose and execute the task. 
In the process, the framework performed multiple rounds of interactive data exploration, dataset partitioning, model design, training, and prediction. 
During execution, several code errors occurred; these were resolved through code repair module which is implemented by transitions between self-debugging and post-filtering, with the framework effectively consolidating past mistakes into the final context history. 
This example highlights the efficacy and flexibility of the FST-based multi-stage architecture in unified interaction representation, which leverages the reasoning and coding capabilities of large language models alongside dynamic environmental interactions to accomplish complex and multifaceted data science tasks.

\section{Details of DFST-based Multi-Stage Architecture}
\label{sec:appendix-fst}

\subsection{State Transition of DFST and Action Signal Space} 
\label{subsec:appendix-state_transition_action_signal}
Built on a finite state transducer (FST), \framework orchestrates four distinct stages—DFS-like planning, incremental execution, self-debugging, and post-filtering.
At each stage, \framework samples an action signal from the predefined action signal space (see \cref{tab:appendix-state_action_signal}), which participates in driving the state transition while triggering the generation and execution of the corresponding markdown and code cells.
We model this multi-stage architecture as a deterministic FST as illustrated in Figure~\ref{fig:appendix_state_transition}.
In the event of an execution error during either the DFS-like planning or incremental execution stage, \framework transitions to the self-debugging and post-filtering stage for code repair.
After post-filtering, the flow returns to the subsequent stage that would normally follow if no error had occurred.

\subsection{Prompts for Each Stage}
\label{subsec:prompt_for_each_stage}
To give readers a clearer understanding of the agent’s behavior at each stage, we detail the prompts used by the agent to generate actions in different states in ~\cref{fig:initial_prompt,fig:planning_prompt,fig:execution_prompt,fig:debugging_prompt,fig:filtering_prompt}.

\subsection{Implementation Details} 
\label{subsec:appendix-implementation_details}

To prevent the FST from entering an infinite loop, we count and limit the number of transitions across three stages: DFS-like planning, incremental execution, and self-debugging. 
We introduce the following hyperparameters: (1) $\texttt{max\_planning\_number}$: the maximum number of transitions into the DFS-like Planning stage; 
(2) $\texttt{max\_execution\_number}$: the maximum number of transitions into the Incremental Execution stage for a given subtask; 
(3) $\texttt{max\_debug\_number}$: the maximum number of consecutive transitions into the Self-Debugging stage during code repair, representing the upper bound on debugging attempts for a single error; 
and (4) $\texttt{max\_planning\_execution\_number}$: the maximum number of non-root nodes in the agent trajectory tree, where actions from both the DFS-like Planning and Incremental Execution phases are considered as nodes. 
Notably, $\texttt{max\_planning\_execution\_number}$ serves to constrain the overall search cost of the solution space.
Those hyperparameters are uniformly configured in our experiments.

\section{Experiments Details}
\label{sec:appendix-experiment-details}

\subsection{Datasets}
\label{subsec:appendix-datasets}
\paragraph{InfiAgent-DABench.} 
We use InfiAgent-DABench~\cite{hu2024infiagent}, a benchmark specifically designed to evaluate agent performance on data analysis tasks. It comprises 257 real-world challenges, each accompanied by a CSV input file and one or more questions related to the data. The challenges span various categories such as summary statistics, feature engineering, and correlation analysis, and are labeled with one of three difficulty levels: easy, medium, or hard.

\paragraph{MatplotBench.}
We adopt MatplotBench~\cite{yang-etal-2024-matplotagent}, a benchmark for the automatic and quantitative evaluation of AI methods in scientific data visualization. It contains 100 curated test cases, each consisting of a user query, an associated input dataset, and a ground-truth figure verified by human experts. The benchmark enables rigorous assessment of plotting accuracy and visual reasoning capabilities.

\paragraph{DSBench.}
We further utilize the data modeling part from DSBench~\cite{jing2024dsbench}, designed to assess agents on complex, real-world data science problems. DSBench includes 74 predictive modeling tasks derived from competitive platforms such as ModelOff\footnote{\url{https://corporatefinanceinstitute.com/resources/financial-modeling/modeloff-guide/}} and Kaggle\footnote{\url{https://www.kaggle.com/}}. Each task provides a large-scale training/test dataset, sample submission file, and a detailed problem description, requiring agents to build end-to-end modeling solutions.

\subsection{Metrics}
\label{subsec:appendix-benchmark_metrics}

\paragraph{InfiAgent-DABench.} 
InfiAgent-DABench~\cite{hu2024infiagent} comprises 258 challenges, each paired with a corresponding CSV input file. These challenges are categorized into three difficulty levels (easy, medium, or hard) and include one or more questions about the data. For close-form questions, we define the following metrics:

\begin{itemize}

\item \textbf{Proportional Accuracy by Subquestions (PASQ)}:
\begin{equation}
\text{PASQ} = \frac{1}{N} \sum_{i=1}^{N} \left( \frac{1}{M_i} \sum_{j=1}^{M_i} I_{ij} \right)
\end{equation}
Here, $N$ denotes the total number of questions, $M_i$ is the number of subquestions in the \(i\)-th question, and $I_{ij}$ is the indicator function for the \(j\)-th subquestion of the \(i\)-th question.

\item \textbf{Accuracy by Questions (ABQ)}
\begin{equation}
\text{ABQ} = \frac{1}{N} \sum_{i=1}^{N} \left( \prod_{j=1}^{M_i} I_{ij} \right)
\end{equation}
The product $\prod_{j=1}^{M_i} I_{ij}$ equals 1 if all subquestions of the \(i\)-th question are answered correctly, and 0 otherwise.

\item \textbf{Uniform Accuracy by Subquestions (UASQ)}
\begin{equation}
\text{UASQ} = \frac{1}{\sum_{i=1}^{N} M_i} \sum_{i=1}^{N} \sum_{j=1}^{M_i} I_{ij}
\end{equation}

\end{itemize}

\paragraph{DSBench.}
The 74 data modeling tasks from DSBench~\cite{jing2024dsbench} are sourced from real-world Kaggle competitions and feature large-scale training and testing datasets along with complex instructions, making them particularly challenging for data science agents.
For evaluation, DSBench first adopts \textbf{Task Success Rate}, which measures whether the agent successfully builds a machine learning model and generates a bug-free submission.
However, due to the inconsistency of metric scales and evaluation dimensions across different tasks, directly comparing performance is non-trivial. To address this, DSBench introduces the \textbf{Relative Performance Gap (RPG)} as an additional metric to normalize results across diverse tasks. RPG measures the agent’s relative improvement over a baseline, scaled by the gap between the baseline and the best-known performance, and is defined as:
\begin{equation}
\text{RPG} = \frac{1}{N} \sum_{i=1}^{N} \max\left( \frac{p_i - b_i}{g_i - b_i}, 0 \right)
\end{equation}
where $N$ is the total number of competitions, $p_i$ is the performance of the agent’s submission for the $i$-th competition, $g_i$ is the highest known performance for the $i$-th competition, and $b_i$ is the performance of a baseline. DSBench~\cite{jing2024dsbench} uses the performance of the original submission file in the competition as the baseline
performance in the RPG computation process.

\subsection{\framework and Baselines Configurations} 
\label{subsec:appendix-configurations}
\paragraph{\framework Configuration.} 
For the experiments on \textbf{data analysis} and \textbf{scientific visualization}, we set $\texttt{max\_planning\_number}=7$, $\texttt{max\_execution\_number}=6$, and $\texttt{max\_debug\_number}=8$. 
For \textbf{predictive modeling}, we configured the hyperparameters as $\texttt{max\_planning\_number}=7$, $\texttt{max\_execution\_number}=6$, $\texttt{max\_debug\_number}=8$, and $\texttt{max\_planning\_execution\_number}=15$. 
These predefined hyperparameters act as guardrails to ensure robust performance and maintain a consistent experimental environment. They do not constrain the generality of our approach but rather provide necessary safeguards against unexpected failures. 

\begin{table}[htbp]
\centering
\small
\begin{tabular}{cl}
\toprule
Avg. LLM calls & Benchmark \\
\midrule
6.42 & InfiAgent-DABench \\
6.41 & MatplotBench \\
7.56 & MatplotBench(\textbf{w/ visual tool}) \\
12.31 & Data Modeling \\
\bottomrule
\end{tabular}
\caption{Average number of LLM calls across benchmarks made by \framework using GPT-4o.}
\label{tab:appendix_llm_calls}
\end{table}
\begin{table*}[htbp]
    \renewcommand{\arraystretch}{0.9}  
    \centering
    \small
    \resizebox{\linewidth}{!}{
    \begin{tabular}{ll|ccc}
        \toprule
        \textbf{Model} & \textbf{Framework} & \textbf{PASQ/\% $\uparrow$} & \textbf{ABQ/\% $\uparrow$} & \textbf{UASQ/\% $\uparrow$} \\
        \midrule
        \multirow{5}{*}{GPT-4o mini} & ReAct & \underline{85.30} & \underline{80.08} & \underline{84.55} \\
        & AutoGen & 74.68 & 70.04 & 77.41 \\
        & Taskweaver & 81.95 & 76.65 & 81.34 \\
        & Data Interpreter & 73.85 & 67.7 & 72.15 \\
        & \textbf{\framework(Ours)} & \textbf{88.39} & \textbf{82.88} & \textbf{87.06} \\
        \midrule
        \multirow{6}{*}{GPT-4o} & ReAct & 87.48 & 81.32 & 86.62 \\
        & AutoGen & 76.43 & 73.54 & 79.39 \\
        & Taskweaver & \underline{89.35} & \underline{85.99} & \textbf{90.24} \\
        & \textcolor{gray}{Data Interpreter}$^{*}$ & \textcolor{gray}{-} & \textcolor{gray}{\textbf{94.93}} & \textcolor{gray}{-} \\
        & Data Interpreter & 79.97 & 75.78 & 79.59 \\
        & \textbf{\framework(Ours)} & \textbf{89.95} & \underline{85.99} & \underline{89.91} \\
        \midrule
        \multirow{4}{*}{Qwen2.5-72B-Instruct} & ReAct & \underline{82.39} & \underline{75.88} & \underline{78.73} \\
        & AutoGen & 73.87 & 70.04 & 75.22 \\
        & \textbf{\framework(Ours)} & \textbf{87.27} & \textbf{81.71} & \textbf{85.09} \\
        \bottomrule
    \end{tabular}
    }
    \vspace{-0.5em}
    \caption{\textbf{Performance comparison on InfiAgent-DABench across various model settings.} The result marked with an asterisk ($^{*}$) is reported by \citet{hong2024data}. Best results are in bold; second-best are underlined.}
    \label{tab:full_infiagent}
\end{table*}
\begin{table*}[htbp]
    \renewcommand{\arraystretch}{0.86}  
    \centering
    \small  
    \resizebox{\linewidth}{!}{
    \begin{tabular}{ll|cccc}
        \toprule
        \textbf{Model} & \textbf{Framework} & \textbf{Comp. Rate/\%} & \textbf{Scores $\geq$ 80/\%}  & \textbf{Avg. Score$\uparrow$} & \textbf{$\Delta$ Avg. Score}\\
        \midrule
        \multirow{6}{*}{GPT-4o mini} & Direct Decoding & 61 & 24 & 38.09 & -\\
        & MatplotAgent & \underline{94} & 33 & 51.44 & +13.35\\
        & AutoGen & 92 & 32 & 51.82 & +13.73\\
        & \quad w/ visual tool & 90 & 32 & 52.07 & +13.98\\
        & \textbf{\framework(Ours)} & \textbf{99} & \underline{34} & \underline{55.85} & \underline{+17.76}\\
        & \textbf{\quad w/ visual tool} & \textbf{99} & \textbf{39} & \textbf{58.60} & \textbf{+20.51}\\
        \midrule
        \multirow{6}{*}{GPT-4o} & Direct Decoding & 68 & 32 & 45.28 & -\\
        & MatplotAgent & 95 & 41 & 57.86 & +12.58\\
        & AutoGen & 97 & 39 & 60.42 & +15.14\\
        & \quad w/ visual tool & \underline{99} & 36 & \underline{63.60} & \underline{+18.32}\\
        & \textbf{\framework(Ours)} & \textbf{100} & \underline{43} & 61.22 & +15.94\\
        & \textbf{\quad w/ visual tool} & \underline{99} & \textbf{44} & \textbf{64.33} & \textbf{+19.05}\\
        \midrule
        \multirow{5}{*}{\makecell[l]{Qwen2.5-72B\\-Instruct}} & Direct Decoding & 73 & 35 & 47.54 & -\\
        & AutoGen & 65 & 26 & 40.80 & -6.74\\
        & \quad w/ visual tool & 85 & 32 & 53.72 & +6.18\\
        & \textbf{\framework(Ours)} & \underline{98} & \underline{37} & \underline{56.41} & \underline{+8.87}\\
        & \textbf{\quad w/ visual tool} & \textbf{99} & \textbf{42} & \textbf{61.88} & \textbf{+14.34}\\
        \bottomrule
    \end{tabular}
    }
    \vspace{-0.7em}
    \caption{\textbf{Performance comparison on MatplotBench.} We report three metrics: \textbf{Completion Rate} (valid output rate), \textbf{Scores $\geq$ 80} (proportion of high-quality completions), and \textbf{Average Score} (0–100). The last column ($\Delta$ Avg. Score) denotes the score gain over Direct Decoding. Bold and underline highlight the best and second-best results, respectively. Visual tool rows indicate integration with the GPT-4o mini-based visual tool.}
    \label{tab:full_matplotbench}
\end{table*}

To investigate the degree to which \framework adheres to the designed state machine during the experiments, we recorded the average number of LLM calls made by \framework with GPT-4o. 
The results, as illustrated in ~\cref{tab:appendix_llm_calls}, indicate that \framework, through its FST-based multi-stage architecture, effectively orchestrates the transitions among the four key stages.

\paragraph{Details of Experimental Setups.} 
(1) \textbf{Data Analysis.} We benchmark \framework in InfiAgent-DABench against several state-of-the-art agent systems (SoTA), including ReAct\cite{hu2024infiagent}, AutoGen\cite{wu2023autogen}, Taskweaver\cite{qiao2023taskweaver} and Data Interpreter\cite{hong2024data}. 
For model configuration, we set the temperature to 0 for all agents, except for ReAct, where the temperature is set to 0.2, as required by \citealp{hu2024infiagent}.

(2) \textbf{Scientific Visualization.} We benchmark \framework against three baselines in different model settings: Direct Decoding, MatplotAgent, and AutoGen. MatplotBench employs a vision-based scoring mechanism aligned with human assessment \cite{yang-etal-2024-matplotagent}, where an advanced multi-modal LLM, such as GPT-4V\cite{achiam2023gpt}, is prompted to score the generated figure on a scale from 0 to 100, comparing it with the ground truth figure.
Since OpenAI deprecated GPT-4V during our experiments, we adopt GPT-4o, a more powerful version with enhanced vision capabilities, as the recommended replacement by OpenAI\cite{hurst2024gpt}, to serve as the scoring model. The temperature is set to 0 in all methods. 

(3) \textbf{Predictive Modeling.} We evaluate \framework using the experimental setup described in DSBench \cite{jing2024dsbench} and compare its performance with the results reported for AutoGen \cite{wu2023autogen} and Code Interpreter\footnote{\url{https://platform.openai.com/docs/assistants/tools/code-interpreter}}. 
The primary metrics include \textbf{Task Success Rate}, which measures whether the data science agent successfully completes the predictive task, and the \textbf{Relative Performance Gap (RPG)}, which quantifies the overall performance of a data science agent across different competitions. 
We also record \textbf{Inference Time}, the average time taken to complete a task. Each task is assigned a maximum time limit of 3600 seconds, as some competitions involve large datasets or complex tasks that could require extensive computation time. If the task exceeds this limit, it is marked as incomplete, and the time is recorded as 3600 seconds.
The temperature of \framework is set to 0. 

In~\citealt{jing2024dsbench}, detailed specifications of the experimental environment are not provided, and it is challenging to control for resources and environmental factors across different methods. 
Moreover, since inference time is affected by numerous factors, comparing the inference time of \framework with that of AutoGen and Code Interpreter may not yield meaningful insights.
Nevertheless, for the experiments of \framework on the data modeling tasks from DSBench, we conducted all evaluations under a consistent environment to ensure fairness and reproducibility. 
Specifically, the experiments were run on a machine with 80 CPU cores, 512 GB of RAM. 
The operating system is Ubuntu 24.04.1 LTS, and the software environment is managed via Conda with Python 3.10. Core libraries for predictive modeling, such as NumPy (v2.2.1), Pandas (v2.2.3), Matplotlib (v3.10.0), SciPy (v1.15.1), Scikit-learn (v1.6.1), and PyTorch (v2.5.1+cu121), were pre-installed to support the experiments. Additionally, \framework is capable of dynamically installing required packages during task execution by executing command-line installation commands within code cells.

\subsection{Visual Tool for Scientific Visualization} 
\label{subsec:appendix-visual_tool}
We implement a visual tool based on GPT-4o mini to evaluate \framework’s capability in completing scientific visualization tasks through the integration of visual feedback tools. In our experiments, each test case can call the visual tool at most \textbf{four} times for both AutoGen and \framework. Figure~\ref{fig:code_visual_tool} illustrates the details of the implementation and integration of this tool.

\subsection{Full Experimental Results on Data Analysis, Scientific Visualization, and Predictive Modeling} 
\label{subsec:appendix-full_results}
In addition to the main results reported in~\cref{sec:experiments}, we present the complete experimental results for \textbf{Data Analysis}, \textbf{Scientific Visualization}, and \textbf{Predictive Modeling} in Tables~\cref{tab:full_infiagent},~\cref{tab:full_matplotbench}, and~\cref{tab:full_datamodeling}, respectively. These tables include additional evaluation metrics (with definitions provided in~\cref{subsec:appendix-benchmark_metrics}), offering a more comprehensive assessment of \framework’s performance across different tasks. As shown in~\cref{tab:full_matplotbench}, \framework achieves substantially higher scores than all baselines in both \emph{Completion Rate} and the \emph{Proportion of Scores} $\geq$ 80. These results highlight \framework’s robustness in task completion as well as its ability to generate consistently high-quality visual outputs.

\begin{figure*}[htbp]
\begin{tcolorbox}[colback=blue!5!white,colframe=blue!75!black,title=PLAIN TEXT, halign title=center]
\begin{lstlisting}[
  breaklines=true,
  breakatwhitespace=false,
  basicstyle=\ttfamily\scriptsize,   % 更小字体
  aboveskip=0.5em,                   % 上方间距缩小
  belowskip=0.5em,                   % 下方间距缩小
  % lineskip=-1pt                      % 行间更紧凑（可选）
]
I have a data modeling task. You must give me the predicted results as a CSV file as detailed in the following content. You should try your best to
predict the answer. I provide you with three files. One is training data, one is test data. There is also a sample file for submission.
Description
The May edition of the 2022 Tabular Playground series binary classification problem includes a number of different feature interactions. This
competition is an opportunity to explore various methods for identifying and exploiting these feature interactions.
About the Tabular Playground Series
Kaggle competitions are incredibly fun and rewarding, but they can also be intimidating for people who are relatively new to their data science journey.
In the past, we've launched many Playground competitions that are more approachable than our Featured competitions and thus, more beginner-
friendly.
The goal of these competitions is to provide a fun and approachable-for-anyone tabular dataset to model. These competitions are a great choice for
people looking for something in between the Titanic Getting Started competition and the Featured competitions. If you're an established competitions
master or grandmaster, these probably won't be much of a challenge for you; thus, we encourage you to avoid saturating the leaderboard.
For each monthly competition, we'll be offering Kaggle Merchandise for the top three teams. And finally, because we want these competitions to be
more about learning, we're limiting team sizes to 3 individuals.
Getting Started
For ideas on how to improve your score, check out the Intro to Machine Learning and Intermediate Machine Learning courses on Kaggle Learn.
We've also built a starter notebook for you that uses TensorFlow Decision Forests, a TensorFlow library that matches the power of XGBoost with a
friendly, straightforward user interface.
Good luck and have fun!
Acknowledgments
Photo by Clarisse Croset on Unsplash.
Evaluation
Submissions are evaluated on the area under the ROC curve between the predicted probability and the observed target.
Submission File
For each id in the test set, you must predict a probability for the target variable. The file should contain a header and have the following format:
```
id, target
900000, 0.65
900001, 0.97
900002, 0.02
etc.
```
Dataset Description
For this challenge, you are given (simulated) manufacturing control data and are tasked to predict whether the machine is in state 0 or state 1. The
data has various feature interactions that may be important in determining the machine state. Good luck!
Files
- train.csv: the training data, which includes normalized continuous data and categorical data
- test.csv: the test set; your task is to predict the binary target variable which represents the state of a manufacturing process
- sample_submission.csv: a sample submission file in the correct format
All three data files can be found in the folder `./input/`. **You should use sklearn or pytorch to complete the task.** Any training scripts, models, and
experiment log should be saved in `./input/`.
After data modeling, provide the prediction results for the test file in the format specified by the sample submission file. Save the final submission to
`./input/final_submission.csv`.
\end{lstlisting}
\end{tcolorbox}
\caption{The complete instruction of the data modeling task with index = 48.} 
\label{fig:instruction_case_example_data_modeling}  
\end{figure*}

\begin{figure*}[htbp]
\begin{tcolorbox}[colback=blue!5!white,colframe=blue!75!black,title=PSEUDOCODE, halign title=center]
\begin{lstlisting}[
    breaklines=true,
    breakatwhitespace=false,
    basicstyle=\ttfamily\small,   % 更小字体
    aboveskip=0.5em,                   % 上方间距缩小
    belowskip=0.5em,                   % 下方间距缩小
]
GLOBAL_CNT <- 4
EVALUATION_CNT <- 0

function evaluate_image(image_path, requirements, query):
    if EVALUATION_CNT >= GLOBAL_CNT:
        return "Usage limit reached. Please manually evaluate."

    if image_path is invalid or does not exist:
        raise error

    if requirements or query is empty:
        raise error

    encoded_image <- encode_image_to_base64(image_path)

    prompt <- "Expected Requirements:\n" + requirements
    prompt += "\nQuery:\n" + query
    prompt += "\nYour response:\n"

    message <- [
        {"type": "text", "text": prompt},
        {"type": "image_url", "image_url": {"url": encoded_image}}
    ]

    try:
        response <- call_chat_completion(model="gpt-4o-mini", message)
        EVALUATION_CNT += 1
        return response.content
    except:
        raise runtime_error
\end{lstlisting}
\end{tcolorbox}
\caption{Pseudocode of the GPT-4o mini-based visual tool. This tool generates a textual response to a given query by analyzing the provided image in light of the specified requirements.}
\label{fig:code_visual_tool}
\end{figure*}

\begin{figure*}[htbp]
\begin{adjustbox}{max width=\linewidth, max totalheight=0.9\textheight}
\begin{tcolorbox}[colback=blue!5!white,colframe=blue!75!black,title=PROMPT, halign title=center]
\begin{lstlisting}[
    breaklines=true,
    breakatwhitespace=false,
    basicstyle=\ttfamily\small,   % 更小字体
    aboveskip=0.5em,                   % 上方间距缩小
    belowskip=0.5em,                   % 下方间距缩小
]
The current [USER INSTRUCTION]:
{{the description of user instruction}}

The current [STEP GOAL]:
[STEP GOAL]: {{the description of current step}}

The current STEP has been finished as above. Currently in the Planning Stage.
Available Action Space: {<Iterate on Current STEP>, <Advance to Next STEP>, <Fulfill USER INSTRUCTION>}

Your response MUST start with **exactly one** of the action signals, and then generate the corresponding action:

1. `<Iterate on Current STEP>`: 
    When to choose: Select this if the current STEP was incorrect or requires replacement.
    Action content: Write observations from the wrong current STEP when needed. Then reinitiate a NEW AND DISTINCT [STEP GOAL] and write cells incrementally for kernel execution to REPLACE the current one.
    Response Format: 
    <Iterate on Current STEP>
    ```markdown
    (observations in detailed from the replaced STEP when needed)
    ```
    ```markdown
    [STEP GOAL]: (the description of [STEP GOAL])
    ```
    ```markdown/python
    # several markdown and code cells
    ```

2. `<Advance to Next STEP>`: 
    When to choose: Select this when the current STEP was successful and provides a correct foundation for further progress.
    Action content: Define the next [STEP GOAL] to **progress towards fulfilling the [USER INSTRUCTION]**. Write cells to implement the next step incrementally.

    One [STEP GOAL] could be initiated in one markdown cell labeled with `[STEP GOAL]: `.

3. `<Fulfill USER INSTRUCTION>`:
    When to choose: Select this if the current [USER INSTRUCTION] has been fully satisfied, and no further STEPs are necessary.
    Action content: Conclude the process by providing a thorough and structured summary that encapsulates all key aspects of the completed [USER INSTRUCTION]. The summary should be clear, concise, and organized to ensure the user fully understands the results and implications of the task.

Selecting <Iterate on Current STEP> or <Advance to Next STEP> transitions the workflow to the Incremental Execution Stage to implement the new or next STEP, while selecting <Fulfill USER INSTRUCTION> concludes the workflow and no further stages are required.

Specifically, the response format is below:

<Iterate on Current STEP>/<Advance to Next STEP>/<Fulfill USER INSTRUCTION>
```markdown/python
# several markdown and code cells
```

Your response:
\end{lstlisting}
\end{tcolorbox}
\end{adjustbox}
\caption{The prompt of the DFS-like planning stage.} 
\label{fig:planning_prompt} 
\end{figure*}

\begin{figure*}[htbp]
\begin{tcolorbox}[colback=blue!5!white,colframe=blue!75!black,title=PROMPT, halign title=center]
\begin{lstlisting}[breaklines=true, breakatwhitespace=false]
The current [STEP GOAL]:
[STEP GOAL]: {{the description of current step}}
Currently in the Incremental Execution Stage.
Available Action Space: {<await>, <end_step>}
Your response MUST start with **exactly one** of the action signals and then generate the corresponding action:
1. `<await>`:
When to choose: Select this when you need code to be executed in the Jupyter kernel.
Action content: Writing several markdown and code cells.
2. `<end_step>`:
When to choose: Select this when the current [STEP GOAL] has been fully completed.
Action content: Indicate that the step is finished and write any final cells needed to finalize the STEP.
Selecting `<end_step>` transitions the workflow to the **Planning Stage** to evaluate the next step or finalize the response to the `[USER
INSTRUCTION]`. Otherwise, stay in the **Incremental Execution Stage** to continue working on the current STEP.
Specifically, the response format is below:
<await>/<end_step>
```markdown/python
# several markdown and code cells
```
Your response:
\end{lstlisting}
\end{tcolorbox}
\caption{The prompt in the incremental execution stage.}
\label{fig:execution_prompt}
\end{figure*}

\begin{figure*}[htbp]
\begin{tcolorbox}[colback=blue!5!white,colframe=blue!75!black,title=PROMPT, halign title=center]
\begin{lstlisting}[breaklines=true, breakatwhitespace=false]
The current [STEP GOAL]:
[STEP GOAL]: {{the description of current step}}

Currently in the Debugging Stage.
Available Action Space: {<await>, <end_debug>}

Your response MUST start with **exactly one** of the action signals, and then generate the corresponding action:

1. `<await>`: 
    When to choose: Select this when you need code cells to be executed in the Jupyter kernel to diagnose issues, test fixes, or inspect intermediate results.
    Action content: Writing several markdown and code cells containing your debugging attempts, such as code modifications, print statements, or variable inspections.
2. `<end_debug>`: 
    When to choose: Select this when **all bugs are fixed** and you are ready to exit the debugging phase and advance towards the `[STEP GOAL]`.
    Action content: Indicate that debugging is complete and write any final cells needed.

Selecting `<end_debug>` transitions the workflow to the **Post-Debugging Stage** to evaluate and clean up the debugging artifacts. Otherwise, stay in the **Debugging Stage** to continue diagnosing and fixing issues.

Specifically, the response format is below:
<await>/<end_debug>
```markdown/python
# several markdown and code cells
```

Your response:
\end{lstlisting}
\end{tcolorbox}
\caption{The prompt in the self-debugging stage.} 
\label{fig:debugging_prompt}
\end{figure*}

\begin{figure*}[htbp]
\begin{tcolorbox}[colback=blue!5!white,colframe=blue!75!black,title=PROMPT, halign title=center]
\begin{lstlisting}[breaklines=true, breakatwhitespace=false]
Currently in the Post-Debugging Stage.
Available Action Space: {<debug_success>, <debug_failure>}

Your response MUST start with **exactly one** of the action signals, and then generate the corresponding action:

1. `<debug_success>`:
    When to choose: Select this when the debugging phase succeeded in fixing all bugs.
    Action content:
        - Record any valuable information from the debugging process for future reference in a markdown cell.
        - Write one or more **fully cleaned and complete code cells** that **include all necessary steps** to replace the entire debugging process. The provided code must be **self-contained and ready for execution** without requiring any external context or prior cells in the debugging process.
2. `<debug_failure>`:
    When to choose: Select this when the debugging phase failed to fix the bugs.
    Action content: Provide a credible **diagnostic report** based on the debugging process in a markdown cell, explaining what was attempted, why it failed, and any insights from the debugging trace. No code cells should be provided.

Specifically, the response format is below:

<debug_success>/<debug_failure>
```markdown/python
# several markdown and code cells
```

Your response:
\end{lstlisting}
\end{tcolorbox}
\caption{The prompt in the post-filtering stage.}
\label{fig:filtering_prompt}
\end{figure*}

\begin{figure*}[htbp]
\begin{tcolorbox}[colback=blue!5!white,colframe=blue!75!black,title=PROMPT, halign title=center]
\begin{lstlisting}[breaklines=true, breakatwhitespace=false]
Let's address the [USER INSTRUCTION] STEP by STEP. You should initiate ONE [STEP GOAL] first and write markdown and code cells incrementally for kernel execution.

Specifically, the response format is below:

```markdown
[STEP GOAL]: (the description of [STEP GOAL])
```
```markdown/python
# several markdown and code cells
```

Your response:
\end{lstlisting}
\end{tcolorbox}
\caption{The prompt at the start state.} 
\label{fig:initial_prompt}  
\end{figure*}

\end{document}